\documentclass{pinepaper}

\usepackage{multirow}
\usepackage{tabularx}
\usepackage{xspace}
\usepackage{float}
\usepackage{wrapfig}
\usepackage{needspace}

\newcommand{\methodname}{{AffordTrajDP}\xspace}

\title{AffordTrajDP: Dynamic Affordance-Guided Visuomotor Policy Learning for Robotic Manipulation}

\author[1]{Gaoyuan Wu}
\author[1]{Ziyu Shan}
\author[1]{Haoyang Du}
\author[1]{Yuyao Jiang}
\author[1]{Ziwei Wang\textsuperscript{\textdagger}}
\affil[1]{Nanyang Technological University, Singapore}

\date{\small \textsuperscript{\textdagger}Corresponding author.}

\papervenue{Preprint}
\paperdate{\today}
\correspondence{\href{mailto:ziwei.wang@ntu.edu.sg}{ziwei.wang@ntu.edu.sg}}

\begin{document}
\makepinetitle

\begin{pineabstract}
Affordance-guided imitation learning has shown impressive performance in robotic manipulation tasks by compressing visual perception into task-specific geometric constraints (e.g., fixed contact points). However, the commonly used static affordances can become inconsistent in precision-critical tasks or under object location perturbations, leading to post-contact trajectory drift. To address this issue, we propose AffordTrajDP, a dynamic framework that constructs affordance trajectories via object-centric temporal propagation to guide the progressive manipulation process. Specifically, given an RGB-D observation, our core insight is that a retrieved anchor affordance, which captures the desired contact point between the end-effector and the target object, can be propagated forward via affordance propagation, using the object's SE(3) pose as a natural propagation medium, to yield an affordance trajectory that provides temporally consistent, state-aware guidance throughout execution. AffordTrajDP achieves 70.0\% average success rate on ManiSkill3, outperforming strong baselines by up to 17.8\%. Real-world experiments on Galaxea A1 and UR7e robotic arms, covering StackCube, PickCup, AdapterInsertion, Ring-on-Peg, Put-in-Bowl, and USB Insertion, further validate robustness under object placement variations and appearance changes, with seen and unseen object instances evaluated on Galaxea A1, and ablations confirm the contribution of each proposed component.
\end{pineabstract}


\section{Introduction}\label{sec:introduction}

Existing visuomotor policy for robotic manipulation~\cite{chi2025diffusion,geng2022endtoendaffordancelearningrobotic} have attracted increasing attention for their ability to leverage dense visual observations to capture rich scene information and generalize across diverse manipulation tasks. However, this heavy reliance on dense visual inputs makes them sensitive to environmental perturbations such as object appearance and background layout, leading to performance degradation under novel scene configurations and unseen object instances. To mitigate this sensitivity, recent object-centric approaches~\cite{hsu2025spotse3posetrajectory,sun2025prism,yuan2024general,noh20253dflowdiffusionpolicy} adopt structured intermediate representations, such as object pose and 3D point flows, as priors to improve generalization under object position perturbations. By distilling the scene into task-relevant object states and filtering out irrelevant background distractions, these representations generalize robustly across diverse scene configurations, viewpoints, and lighting conditions.

\begin{figure}[t]
  \centering
  \includegraphics[width=\linewidth]{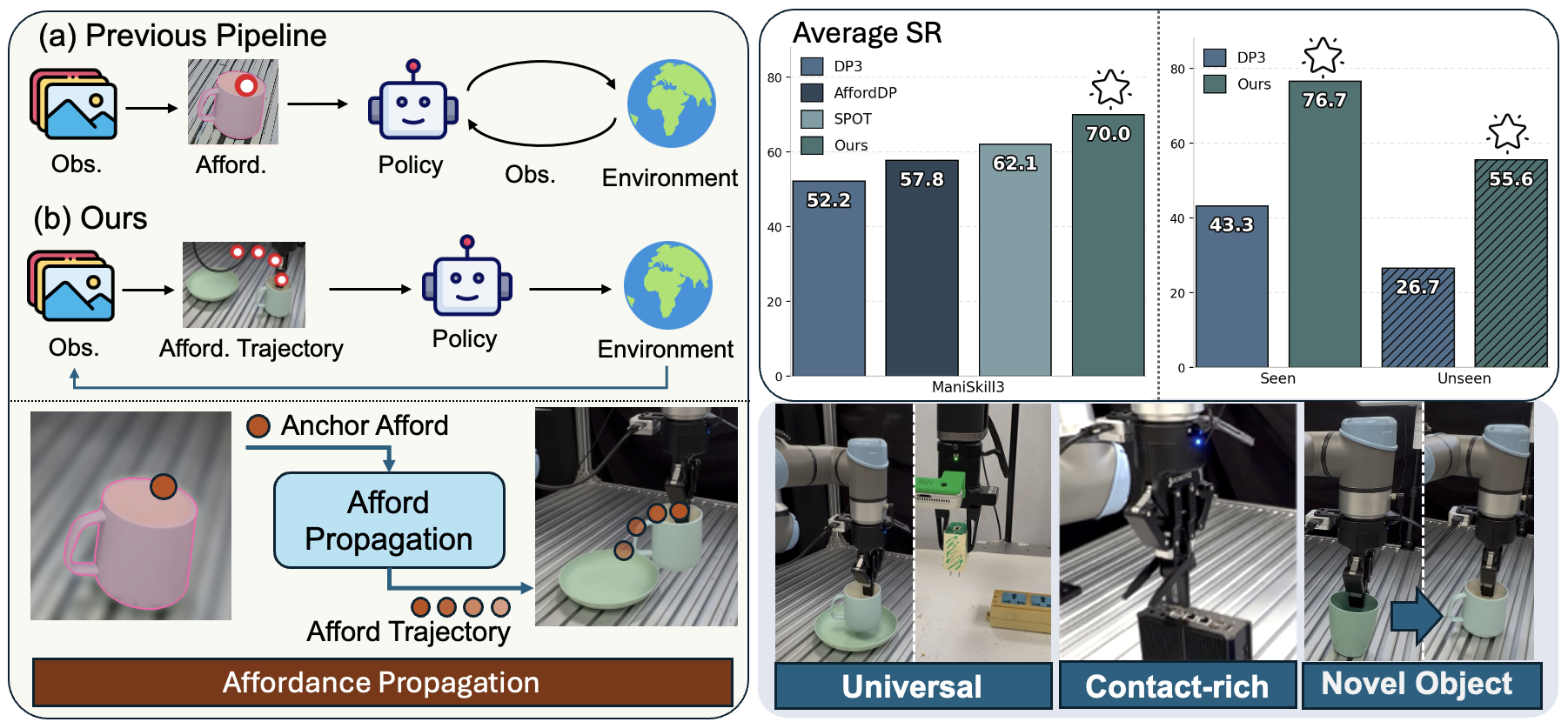}
  \captionsetup{width=\linewidth}
  \caption{\textbf{Overview of \methodname and comparison with previous methods.} (a) Prior pipelines retrieve a static contact point that never updates as the object moves. (b) \methodname propagates it into a time-varying trajectory that continuously conditions the policy, yielding the best ManiSkill3 success rate (70.0\% vs.\ 52.2--62.1\%) and the largest real-world gains (76.7\%/55.6\% vs.\ 43.3\%/26.7\% for DP3 on seen/unseen instances), while generalizing across embodiments, precision insertion, and cluttered scenes.}
  \label{fig:teaser}
\end{figure}

As a compelling form of object-centric representation, affordances encode task-specific interaction priors such as contact locations and post-contact motion. However, existing methods treat these priors as static: AffordDP~\cite{wu2025afforddp} retrieves a contact-point trajectory from a reference demonstration and transfers it to the target scene via a one-shot rigid transformation, after which the guidance remains fixed regardless of contact-induced deviations, so the prior progressively misaligns with the object's evolving pose in precision-critical tasks such as adapter insertion. More broadly, prior conditioning signals capture only half of what is needed: methods that track the object's global pose---whether the current frame (PRISM-DP~\cite{sun2025prism}) or a predicted future trajectory (SPOT~\cite{hsu2025spotse3posetrajectory})---discard the semantics of \emph{where on the object to contact}, while methods that retrieve a semantic contact point (AffordDP) discard \emph{how that contact location should update} as the object moves. Our diagnosis---to our knowledge not made explicit in prior work---is that precision-critical failures stem from the absence of a single signal that jointly preserves both, which motivates both our method and the ablation design in Section~\ref{sec:experiments}. To address this, we propose \textbf{\methodname}, which generates dynamic affordance trajectories via an anchor affordance and an affordance propagation mechanism, providing state-consistent guidance for post-contact motion and improved robustness to environmental perturbations.

We systematically evaluate our method in both simulation and real-world settings.
In ManiSkill3, across six tabletop manipulation tasks, \methodname improves the average success rate from 52.2\% (DP3) to 70.0\% and from 57.8\% (AffordDP) to 70.0\%, demonstrating stronger robustness under diverse task settings.
We further validate \methodname on two real-world robot platforms: a Galaxea A1 arm across three tasks (PickCup, StackCube, and AdapterInsertion), evaluated on both seen and unseen object instances, and a UR7e arm across four tasks (PickCup, Ring-on-Peg, Put-in-Bowl, and USB Insertion). These tasks are representative, spanning a range of manipulation skills from object grasping and object-to-object alignment to precision insertion, enabling a comprehensive evaluation of generalization and robustness.
As shown in Table~\ref{tab:realworld_results}, our method consistently achieves superior success rates on Galaxea A1 and demonstrates impressive generalization and instance-level robustness across unseen instances.
Notably, the improvement is most pronounced on the precision-critical AdapterInsertion task, where static priors and purely point-cloud-based policies often drift after contact, while our dynamic affordance trajectory provides state-consistent guidance that stabilizes post-contact alignment.

In this work, our contributions are threefold:
(1) \textbf{New problem formulation.} We identify and formalize a key limitation of existing affordance-guided policies: static affordance priors, once transferred, cannot maintain alignment with a moving object after contact, causing progressive guidance degradation in precision-critical tasks. We trace this to a missing conditioning signal that jointly preserves \emph{where to contact} and \emph{how that contact should evolve}, and propose \methodname{} as the first framework to address this by constructing a state-consistent, time-varying affordance trajectory as the policy conditioning signal.
(2) \textbf{Necessity and sufficiency of joint conditioning.} Semantic contact retrieval and pose-trajectory prediction have so far only been studied separately; we show via controlled ablation that combining them is both necessary and sufficient. Contact retrieval alone helps but plateaus below our full method (47.2\% vs.\ 35.0\% baseline); trajectory propagation without a semantic anchor is unreliable and can underperform the baseline on unseen instances (StackCube unseen: 20\% vs.\ 40\%); only the combination yields consistent, large gains (66.1\%, Table~\ref{tab:realworld_results}). This is a falsifiable claim about how the two components relate, not just evidence that our method works: each is necessary, neither is sufficient.
(3) \textbf{Empirical validation.} We evaluate \methodname{} on six ManiSkill3 simulation tasks and seven real-world tasks across two robot platforms with seen and unseen object instances, consistently outperforming DP3, AffordDP, and SPOT baselines, with the largest gains on precision-critical, contact-rich tasks.

\section{Related Work}\label{sec:related_work}

\begin{figure}[t]
    \centering
    \includegraphics[width=\linewidth]{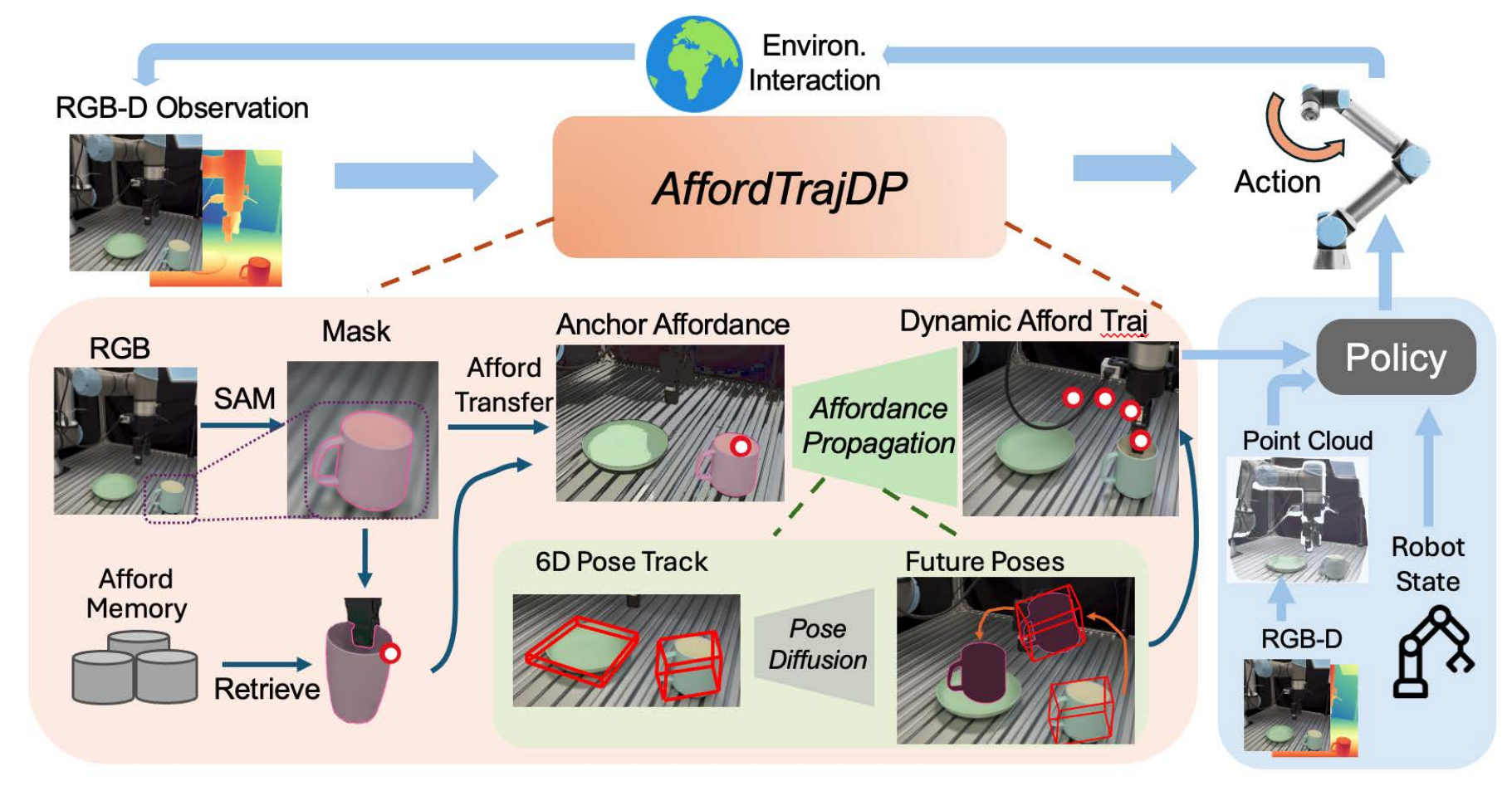}
    \captionsetup{width=\linewidth}
    \caption{\textbf{Pipeline. } Overview. This figure illustrates how AffordTrajDP transfers a static affordance and propagates it into the future to obtain a dynamic affordance trajectory. Given an RGB-D observation, AffordTrajDP first retrieves the most similar object instance from the affordance memory and transfers its static affordance to the target object. In parallel, the pose trajectory generation module takes the current estimated 6D pose as input and predicts a future SE(3) pose trajectory. Then, we generate the dynamic affordance trajectory by propagating the transferred static affordance through the predicted pose trajectory via 6D rigid transformations, producing a dynamic affordance trajectory that conditions the downstream diffusion policy for action generation.}
    \label{fig:pipeline}
\end{figure}

\textbf{Diffusion-based Imitation learning.}
Diffusion Policy~\cite{chi2025diffusion} was among the first to introduce diffusion models into imitation learning, modeling the conditional action distribution via iterative denoising; it offers stable training and multimodal expressivity but generalizes poorly to out-of-distribution settings due to its reliance on raw pixel inputs. Subsequent work~\cite{ke20243ddiffuseractorpolicy,cao2025mambapolicyefficient3d,zhang2026pocketdp3efficientpocketscale3d,deng2020selfsupervised6dobjectpose,reuss2023goalconditionedimitationlearningusing,shan2026dockanywheredataefficientvisuomotorpolicy,shan2026dvgwmdisentangledvideogeneration} improves spatial generalization with 3D representations: 3D Diffusion Policy~\cite{ze20243d} conditions on point clouds for robustness to object position changes; Equibot~\cite{yang2024equibot} adds SIM(3)-equivariant architectures for geometric generalization; 3D flow methods~\cite{yuan2024general,noh20253dflowdiffusionpolicy} predict future 3D point trajectories as an intermediate target for closed-loop control; and GenDP~\cite{wang2024gendp} builds explicit 3D semantic fields from multi-view RGB-D via vision foundation models. However, none of these methods incorporate task-specific contact semantics as policy conditioning, leaving precision-critical post-contact motion under-constrained.

\textbf{Object-Centric Representations for Policy Learning.}
A related line of work replaces high-dimensional image inputs with structured SE(3) object poses as compact policy representations.
PRISM-DP~\cite{sun2025prism} obtains current-frame object poses via segmentation, automatic mesh generation, and pose tracking, feeding them directly to a diffusion policy to enable scalable deployment without manual mesh reconstruction.
SPOT~\cite{hsu2025spotse3posetrajectory} instead predicts future SE(3) object pose trajectories relative to the goal object frame and conditions the policy directly on these predictions, decoupling actions from raw sensory input to support cross-embodiment generalization and learning from action-less demonstrations.

\methodname{} is structurally distinct from both: PRISM-DP conditions on the \emph{current} observed pose, and SPOT conditions on \emph{predicted future poses} directly, whereas in \methodname{} the predicted pose trajectory is only an \emph{intermediate computation}, used to propagate a semantically grounded contact point---retrieved via CLIP-based matching and transferred through dense visual correspondence---into a time-varying affordance trajectory that is the actual conditioning signal. Neither baseline incorporates this contact semantics, the gap diagnosed in Section~\ref{sec:introduction}.

\textbf{Affordance-guided manipulation.}
Object-centric representations improve generalization and interpretability in robotic manipulation~\cite{devin2018deep,wu2023slotdiffusion,wen2022you,zhu2023learning,wu2022vatmartlearningvisualaction}; among them, affordances are particularly appealing as they directly capture the actionable properties of objects.
Early works~\cite{deng20213d,ning2023where2explore,nguyen2023open,yu2025seqaffordsequential3daffordance,tian2025o3affordoneshot3dobjecttoobject,kuang2024ramretrievalbasedaffordancetransfer,gervet2023act3d3dfeaturefield} treat affordances as fixed spatial properties, predicting regions where actions can be applied. AffordDP~\cite{wu2025afforddp} extends this to anchor affordance priors: 3D contact points paired with \emph{demonstration-replayed} pose-contact trajectories, transferred to a new object instance via a one-shot rigid transformation---but once transferred, both remain fixed for the entire episode and cannot adapt to contact-induced deviations or placement perturbations. FSAG~\cite{han2026fsag} extracts fine-grained, temporally aligned grasp affordances from human videos using semantic priors from pretrained diffusion models; AnchorDP3~\cite{zhao2025anchordp3} anchors diffusion predictions to 3D affordances via sparse keypose sequences under vision-language conditioning; Learning from 10 Demos~\cite{rana2025learning10demosgeneralisable} introduces oriented affordance frames that improve spatial and intra-category generalization from as few as 10 demonstrations. Despite their effectiveness, these methods~\cite{nasiriany2024rtaffordanceaffordancesversatileintermediate,li2025learningpreciseaffordancesegocentric} largely treat affordances as static priors---fixed contact points, keyposes, or canonical frames held constant throughout execution. Such priors can become inconsistent under object placement changes and precision-critical interactions, where small post-contact misalignments accumulate and, absent state-consistent, time-varying constraints, often produce drift, oscillation, or failure under occlusion and tracking noise.

\section{Method}\label{sec:method}

\subsection{Preliminaries}

\textbf{Object-centric Diffusion Policy.} Diffusion-based visuomotor policies cast action prediction as conditional generative modeling: rather than regressing a single deterministic action, they learn to sample from the full conditional distribution of expert actions given the current observation. The original Diffusion Policy conditions on raw RGB images, whose high dimensionality and sensitivity to viewpoint, lighting, and background make the learned policy prone to overfitting scene appearance rather than task-relevant geometry. 3D Diffusion Policy (DP3)~\cite{ze20243d} improves generalization by conditioning on point clouds instead of raw images, replacing pixel-level appearance with an explicit, object-centric 3D geometric representation of the scene that is substantially more robust to background clutter, distractor objects, and camera viewpoint shifts. This point-cloud conditioning provides the object-centric substrate on which we later inject affordance priors (Sec.~\ref{sec:propagation}); by itself, however, it captures only scene geometry and offers no explicit signal for where or how the end-effector should contact the target object, motivating the affordance-guided extension below. Starting from Gaussian noise $\mathbf{a}^{K}_t$, the policy iteratively denoises actions conditioned on observation $\mathbf{o}_t$ and proprioception $\mathbf{s}_t$:
\begin{equation}
\mathbf{a}^{k-1}_t = \alpha_k\big(\mathbf{a}^{k}_t - \gamma_k\,\epsilon_\theta(\mathbf{a}^{k}_t,\,k,\,\mathbf{o}_t,\,\mathbf{s}_t)\big) + \sigma_k\,\mathcal{N}(0,\mathbf{I}).
\end{equation}

\textbf{Static Affordance-guided Policy.} Point-cloud conditioning alone leaves the policy to infer contact semantics implicitly from geometry and demonstrations, which is unreliable in precision-critical tasks where the correct interaction region is not the salient geometric feature (e.g., a small handle on a bulky object). To inject this missing signal explicitly, AffordDP~\cite{wu2025afforddp} augments the diffusion policy with an affordance prior $\Phi=(c,\tau)$, where $c\in\mathbb{R}^3$ is a static contact point capturing \emph{where} to interact with the object, and $\tau\in\mathbb{R}^{3\times H}$ is a post-contact trajectory, replayed directly from a reference demonstration, capturing \emph{how} the end-effector should move after contact. Rather than being learned or predicted, both $c$ and $\tau$ are retrieved from a memory of prior demonstrations and transferred from the source instance to the target object through a single rigid-body transformation $\mathbf{T}\in \mathrm{SE}(3)$ estimated between the two instances:
\begin{equation}
c^{tgt} = \mathbf{T}\,\tilde{c}^{src},\qquad \tau^{tgt} = \mathbf{T}\,\tilde{\tau}^{src},
\end{equation}
and the resulting $\Phi^{tgt}=(c^{tgt},\tau^{tgt})$ conditions the policy as a fixed signal throughout execution. This design is effective when the target object's pose does not change appreciably after the transformation is computed, since $\Phi^{tgt}$ is computed once, at the start of the episode, and never updated: it implicitly assumes the demonstrated post-contact trajectory $\tau^{src}$ remains valid under the same rigid transformation that aligned the initial contact point, which holds only if the object does not move relative to the frame in which $\mathbf{T}$ was estimated.

\textbf{Limitation.} As discussed in Sec.~\ref{sec:introduction}, this fixed prior cannot track the object's evolving pose after contact. We address this by propagating the anchor contact point through the object's predicted future pose trajectory to obtain a time-varying affordance (Sec.~\ref{sec:propagation}).

\subsection{Architecture}

Affordances provide task-relevant contact cues for manipulation~\cite{mo2021where2actpixelsactionsarticulated,deng20213daffordancenetbenchmarkvisual,xu2026a0affordanceawarehierarchicalmodel,mazzaglia2024informationdrivenaffordancediscoveryefficient,nguyen2023languageconditionedaffordanceposedetection3d,wu2025ragnet,wu2023learning}. As shown in Fig.~\ref{fig:pipeline}, \methodname (i) retrieves and transfers an anchor contact point from an object memory via semantic correspondence, (ii) predicts the target object's future pose trajectory in SE(3), and (iii) propagates the anchor through the predicted poses to obtain a time-aligned affordance trajectory that conditions a diffusion policy for closed-loop action generation.

\subsection{Anchor Affordance Generation}

We represent 3D affordance as $\Phi = (c, \tau)$, where $c \in \mathbb{R}^3$ denotes a {static} affordance (\textit{i.e.,} a 3D contact point on the object surface), and $\tau \in \mathbb{R}^{3 \times H}$ denotes the corresponding dynamic affordance trajectory as a contact point trajectory over a horizon $H$.

\Needspace{10cm}
\begin{wrapfigure}{r}{0.47\linewidth}
    \vspace{-0.8em}
    \centering
    \includegraphics[width=\linewidth]{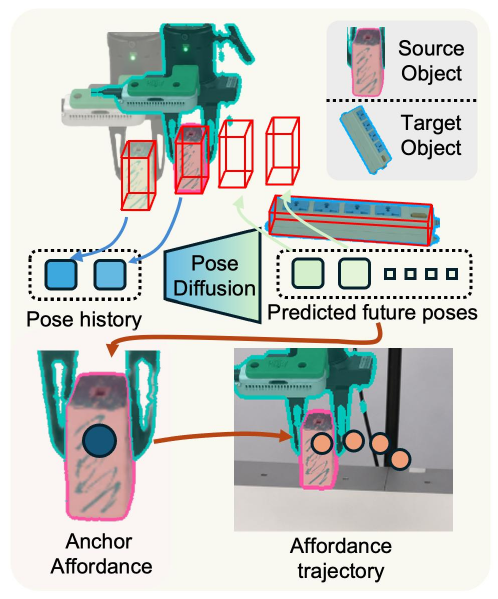}
    \captionsetup{width=\linewidth}
    \caption{\textbf{Dynamic Affordance Trajectory.} A diffusion model predicts future SE(3) poses (green) from the tracked pose history (blue). Propagating the anchor contact point through them, $p_{t+k} = \mathbf{T}_{t+k}\,\tilde{c}^{tgt}$, yields the affordance trajectory $\tau$.}
    \label{fig:vis_afford}
\end{wrapfigure}

We obtain $c$ by retrieval rather than direct regression: the relevant contact region is often a small, sparse part of the object surface that is easily confused with other salient but task-irrelevant geometric features, making it difficult to localize reliably from limited demonstrations without an explicit semantic prior. Retrieval sidesteps this by reducing anchor localization to finding a semantically similar, already-verified source instance and transferring its known contact point via dense correspondence. Note that $c$ alone only specifies \emph{where} to make contact; it carries no information about how the end-effector should move afterward, which motivates propagating it into the trajectory $\tau$ in Sec.~\ref{sec:propagation}.

Given a target object instance, \methodname retrieves the most similar source instance from an object memory built by (i) cropping the object region with GroundedSAM~\cite{ren2024grounded} and (ii) extracting an appearance embedding with the CLIP~\cite{radford2021learning} image encoder. Each memory entry stores the task label $\ell$, static affordance $c$, appearance feature $z$, and RGB-D pose estimate $\mathbf{T}\in SE(3)$:
\begin{equation}
    \mathcal{M} = \{({\ell}, c, z, \mathbf{T})\}.
\end{equation}

At test time, we retrieve the nearest neighbor within the same task $\ell$ by cosine similarity between cropped-image features, obtaining a source contact point $c$. We transfer it to the target via dense visual correspondence between the source and target crops: the target pixel matching the source contact location is back-projected using the target RGB-D observation to yield the target contact point $c^{tgt}\in\mathbb{R}^3$, which serves as the anchor affordance to be propagated along the predicted pose trajectory.

\subsection{Dynamic Affordance Propagation}
\label{sec:propagation}

\textbf{Future Pose Estimation.} For diffusion to be well-defined in Euclidean space, we predict the source object pose expressed in the target object frame, $\mathbf{T}^{src}_{tgt}(t)\in \mathrm{SE}(3)$, rather than in the world frame, which normalizes away absolute scene layout and improves cross-instance generalization.

To train the trajectory diffusion model, we extract object pose trajectories from demonstrations: GroundedSAM~\cite{ren2024grounded} segments the task-relevant object in the first frame via text prompts, Any6D~\cite{lee2025any6d} reconstructs an object mesh from the initial RGB-D view, and FoundationPose~\cite{wen2024foundationpose} tracks the object's 6D pose over time, yielding poses $\mathbf{T}(t)\in \mathrm{SE}(3)$.

To eliminate differences in absolute scene configuration and enable cross-instance learning, we express each demonstrated source-object trajectory in the target object coordinate frame. Letting $\mathbf{T}^{src}_W(t)$ and $\mathbf{T}^{tgt}_W(t)$ denote the world-frame poses of the source and target objects at time $t$, the source pose in the target frame is
\begin{equation}
\mathbf{T}^{src}_{tgt}(t)=\big(\mathbf{T}^{tgt}_W(t)\big)^{-1}\,\mathbf{T}^{src}_W(t),
\label{eq:src_in_tgt}
\end{equation}
which yields an object-centric, layout-invariant trajectory that serves as training supervision for the diffusion model.

\textbf{Affordance Propagation.}
The trajectory diffusion model applies diffusion to the vectorized pose $\mathbf{x}(t)=[\mathbf{t}(t);\mathbf{q}(t)]\in\mathbb{R}^7$, where $\mathbf{t}(t)\in\mathbb{R}^3$ is the translation and $\mathbf{q}(t)\in\mathbb{R}^4$ is a unit quaternion parameterizing $\mathbf{T}(t)\in \mathrm{SE}(3)$, renormalized after each denoising step. The process is conditioned on the pose history $\mathbf{h}_{t}$ rather than raw sensory observations.

Our framework leverages Denoising Diffusion Implicit Models (DDIMs)~\cite{song2020denoising} to model the conditional pose (trajectory) distribution. In contrast to DDPM, DDIM formulates a non-Markovian reverse process whose sampling trajectory is deterministic: for a given conditioning history $\mathbf{h}_t$ and initial noise $\mathbf{x}^{K}_t$, the resulting $\mathbf{x}^{0}_t$ is uniquely determined, precluding the per-step stochastic perturbations that would otherwise propagate into the affordance trajectory (Eq.~\eqref{eq:afford_propagation}). This property further permits accurate sampling with substantially fewer denoising steps, thereby reducing the inference latency of Eq.~\eqref{eq:traj_update} and facilitating real-time control.
\begin{equation}
\mathbf{x}^{k-1}_{t}=\alpha\,\mathbf({x}^{k}_{t}-\gamma\,\epsilon_\theta(\mathbf{h}_{t},\mathbf{x}^{k}_{t},k))+\mathcal{N}(0,\sigma^2\mathbf{I}),
\label{eq:traj_update}
\end{equation}
where the reverse process starts from $\mathbf{x}^{K}_{t}$ sampled from Gaussian noise and iterates for $K$ denoising steps. Correspondingly, the training objective is
\begin{equation}
\mathcal{L}=\mathrm{MSE}\!\big(\epsilon_k,\ \epsilon_\theta(\mathbf{h}_{t},\mathbf{x}^{0}_{t}+\epsilon_k,k)\big).
\label{eq:traj_loss}
\end{equation}

Predicting future object poses turns the static contact prior into a time-varying constraint: the predicted pose trajectory $\{\mathbf{T}_{t+k}\}_{k=0}^{H}$ sampled from this model serves as the temporal backbone along which the anchor contact point $c^{tgt}$ is propagated through time using rigid transformations:
\begin{equation}
\begin{aligned}
    \tau_{t:t+H} &= \big[\,p_t,\ldots,p_{t+H}\,\big],\\
    p_{t+k} &= \mathbf{T}_{t+k}\,\tilde{c},\quad k=0,\ldots,H,
\end{aligned}
\label{eq:afford_propagation}
\end{equation}
where $\tilde{c}=[(c^{tgt})^\top,1]^\top$ is the homogeneous coordinate of $c^{tgt}$, yielding the dynamic affordance trajectory $\tau^{tgt}$ and $\Phi^{tgt}=(c^{tgt},\tau^{tgt})$ (Fig.~\ref{fig:vis_afford}). This propagation produces time-varying contact guidance that adapts to the target object's motion for downstream policy learning.

\subsection{Action Generation}

The static affordance $c$ is encoded by a lightweight MLP~\cite{popescu2009multilayer}, and the dynamic affordance trajectory $\tau$ by a Transformer encoder~\cite{vaswani2017attention} with a prepended $[\mathrm{CLS}]$ token, whose final embedding gives the trajectory feature. Both are combined with other observations to form the conditioning feature $\mathbf{f}_t$.

The action generator is a noise prediction network $\epsilon_\theta$ under the DDIM formulation~\cite{song2020denoising}: starting from Gaussian noise $\mathbf{a}^{K}_t\sim\mathcal{N}(0,\mathbf{I})$, the policy denoises for $K$ steps conditioned on $\mathbf{f}_t$ to obtain a clean action $\mathbf{a}^{0}_t$, via the reverse update
\begin{equation}
\mathbf{a}^{k-1}_t
=
\alpha_k\Big(\mathbf{a}^{k}_t-\gamma_k\,\epsilon_\theta(\mathbf{a}^{k}_t,\,k,\,\mathbf{f}_t)\Big)
+\sigma_k\,\mathcal{N}(0,\mathbf{I}),
\label{eq:action_reverse}
\end{equation}
where $\{\alpha_k\}_{k=1}^{K}$ is the noise schedule and $\gamma_k,\sigma_k$ are step-dependent coefficients. We train $\epsilon_\theta$ to predict the injected noise using the standard denoising objective
\begin{equation}
\mathcal{L}
=
\mathrm{MSE}\!\big(\epsilon^{k},\ \epsilon_\theta(\mathbf{a}^{k}_t,\,k,\,\mathbf{f}_t)\big),
\label{eq:action_loss}
\end{equation}
where $\epsilon^{k}\sim\mathcal{N}(0,\mathbf{I})$.

\section{Experiments}\label{sec:experiments}

\begin{figure}[t]
    \centering
    \includegraphics[width=\linewidth]{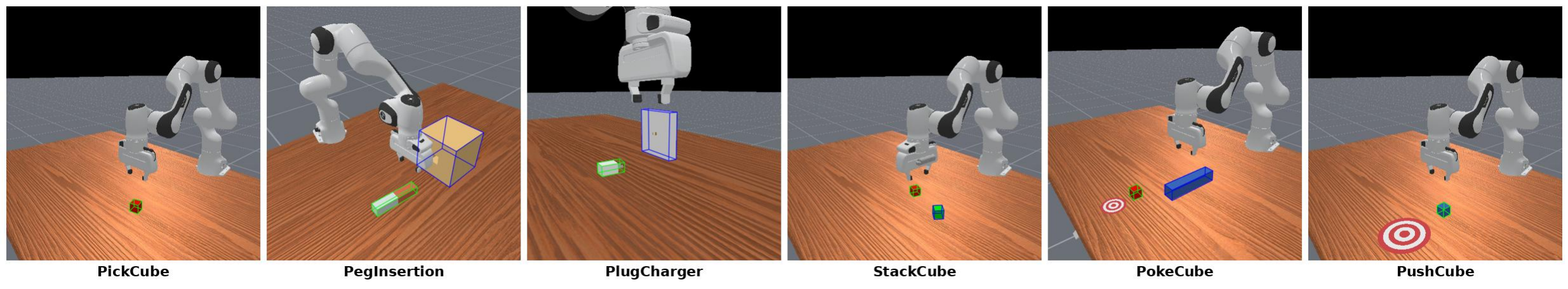}
    \captionsetup{width=\linewidth}
    \caption{\textbf{Simulation tasks.} We evaluate \methodname in the ManiSkill3 simulation environment on six tabletop manipulation tasks: \textit{PickCube}, \textit{PegInsertion}, \textit{PlugCharger}, \textit{StackCube}, \textit{PokeCube}, and \textit{PushCube}. All tasks are executed by a Franka Panda robot with a parallel-jaw gripper.}
    \label{fig:simulation_tasks}
\end{figure}

\begin{table}[t]
\centering
\renewcommand{\arraystretch}{1.15}
\setlength{\tabcolsep}{1pt}
\small
\begin{tabularx}{\linewidth}{l *{6}{>{\centering\arraybackslash}X} >{\centering\arraybackslash}X}
\toprule
\multirow{2}{*}{\textbf{Method}}
  & \multicolumn{6}{c}{\textbf{Tasks Success Rates (\%)\,$\uparrow$}}
  & \multirow{2}{*}{\shortstack{\textbf{Overall SR (\%)}\\$\uparrow$}} \\
\cmidrule(lr){2-7}
 & PickCube & PegInsertion & PlugCharger & StackCube & PokeCube & PushCube & \\
\midrule
DP3      & $46.0 \pm 7.6$          & $23.3 \pm 8.5$          & $11.3 \pm 5.1$          & $72.0 \pm 7.7$           & $76.0 \pm 8.0$ & $84.7 \pm 6.9$ & $52.2$ \\
AffordDP & $52.7 \pm 3.7$          & $36.0 \pm 10.9$         & $14.7 \pm 5.1$          & $75.3 \pm 12.8$          & $80.0 \pm 8.5$ & $88.0 \pm 6.9$ & $57.8$ \\
SPOT     & $56.7 \pm 5.3$          & $40.0 \pm 7.5$          & $12.0 \pm 3.8$          & $84.0 \pm 8.0$           & $86.7 \pm 7.5$ & $93.3 \pm 5.3$ & $62.1$ \\
\rowcolor{gray!25}
Ours     & $\mathbf{60.7} \pm 4.3$ & $\mathbf{46.0} \pm 6.0$ & $\mathbf{24.7} \pm 6.9$ & $\mathbf{92.0} \pm 8.7$  & $\mathbf{96.7} \pm 3.3$ & $\mathbf{100.0} \pm 0.0$ & $\mathbf{70.0}$ \\
\bottomrule
\end{tabularx}
\caption{\textbf{Simulation results on ManiSkill3.} Success rates (\%, mean\,$\pm$\,std over 5 seeds $\times$ 30 trials) on six tabletop manipulation tasks. We use 30 demonstrations for PickCube, StackCube, PokeCube, and PushCube, and 50 for PegInsertion and PlugCharger. All methods share identical demonstrations and hyperparameters. \textbf{Bold} denotes the best result per column.}
\label{tab:sim_results}
\end{table}

\Needspace{11cm}
\subsection{Experimental Setup}

\begin{wrapfigure}{r}{0.55\linewidth}
    \vspace{-0.8em}
    \centering
    \begin{subfigure}[b]{0.49\linewidth}
        \centering
        \includegraphics[height=3.6cm]{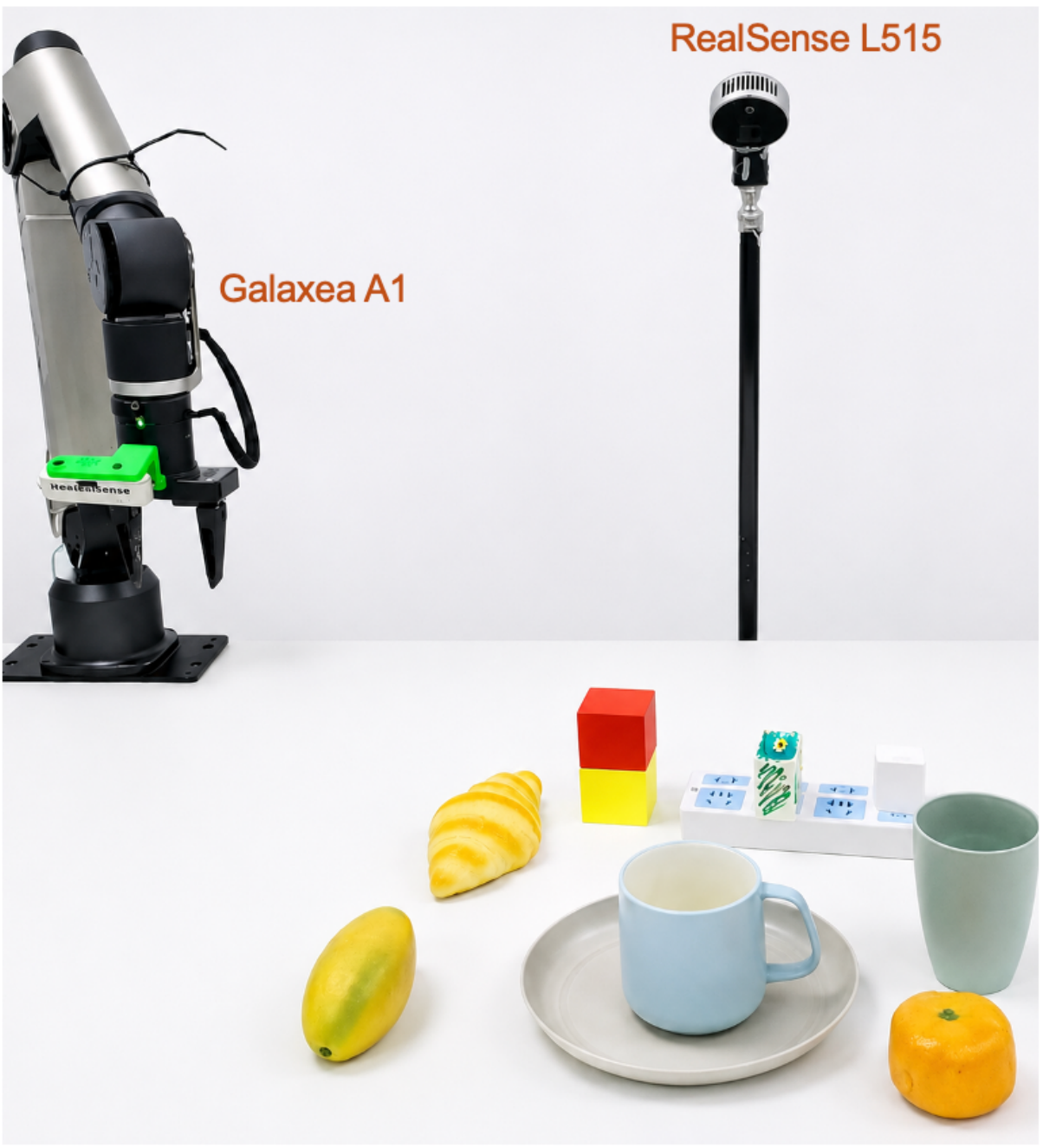}
        \caption{Galaxea A1 + L515}
    \end{subfigure}%
    \begin{subfigure}[b]{0.49\linewidth}
        \centering
        \includegraphics[height=3.6cm]{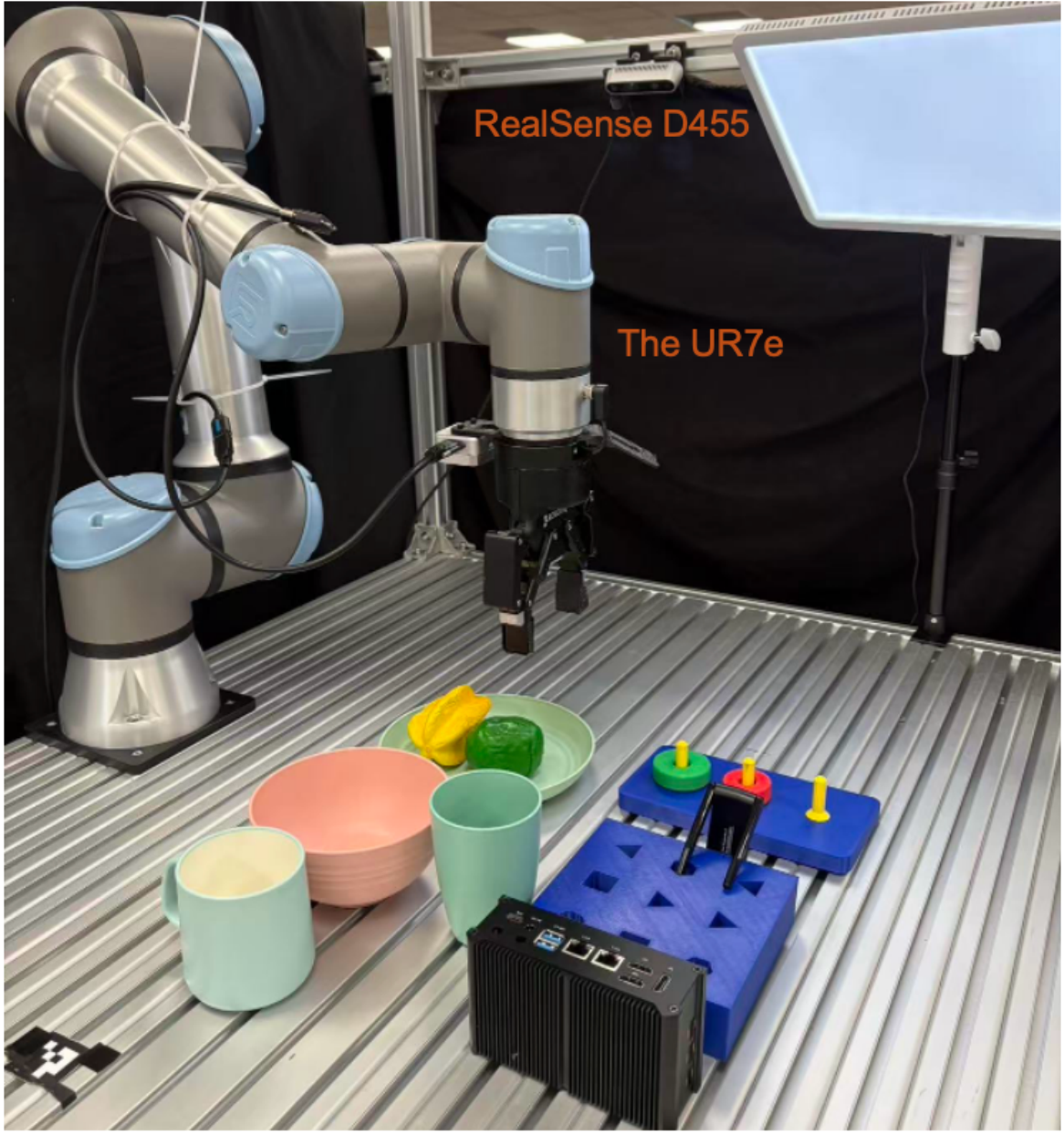}
        \caption{UR7e + D455}
    \end{subfigure}
    \caption{\textbf{Real-world hardware platforms.} We evaluate on two robot platforms, each equipped with a parallel gripper and a third-person RGB-D camera and populated with diverse everyday objects. (a) A Galaxea A1 arm with an Intel RealSense L515 camera, used for \textit{StackCube}, \textit{PickCup}, and \textit{AdapterInsertion}. (b) A UR7e arm with an Intel RealSense D455 camera, used for \textit{PickCup}, \textit{Ring-on-Peg}, \textit{Put-in-Bowl}, and \textit{USB Insertion}.}
    \label{fig:setup}
\end{wrapfigure}

\textbf{Simulation setup.} We evaluate \methodname on six tabletop manipulation tasks in ManiSkill3~\cite{tao2024maniskill3} using a Franka Panda with a parallel-jaw gripper: PickCube, StackCube, PegInsertion, PlugCharger, PokeCube, and PushCube (Fig.~\ref{fig:simulation_tasks}), spanning low- to high-precision contact patterns so we can test whether dynamic affordance guidance helps more as precision demands grow. We report \emph{success rate}, and all methods are trained on the same demonstrations with identical observation and action interfaces.

\textbf{Real-world setup.} As shown in Fig.~\ref{fig:setup}, we use two platforms with a fixed, extrinsically calibrated third-person RGB-D camera: a Galaxea A1 arm (RealSense L515) for StackCube, PickCup, and AdapterInsertion, and a UR7e arm (RealSense D455) for PickCup, Ring-on-Peg, Put-in-Bowl, and USB Insertion. We run 30 trials per method for each object type and report success rates.

\begin{table}[t]
\centering
\renewcommand{\arraystretch}{1.3}
\setlength{\tabcolsep}{6pt}
\small
\begin{tabular}{l cc cc cc c}
\toprule
\multirow{2}{*}{\textbf{Method}} & \multicolumn{2}{c}{AdapterInsertion} & \multicolumn{2}{c}{PickCup} & \multicolumn{2}{c}{StackCube} & \multirow{2}{*}{\shortstack{\textbf{Avg.\ SR (\%)}\\$\uparrow$}} \\
\cmidrule(lr){2-3}\cmidrule(lr){4-5}\cmidrule(lr){6-7}
 & Seen & Unseen & Seen & Unseen & Seen & Unseen & \\
\midrule
DP3      & 3/30  & 0/30  & 20/30 & 12/30 & 16/30 & 12/30 & 35.0 \\
+Center  & 3/30  & 1/30  & 24/30 & 12/30 & 22/30 & 8/30  & 38.9 \\
+Contact & 4/30  & 3/30  & 24/30 & 17/30 & 21/30 & 16/30 & 47.2 \\
+Traj    & 3/30  & 1/30  & 26/30 & 14/30 & 23/30 & 6/30  & 40.6 \\
\rowcolor{gray!25}
Ours     & $\mathbf{12/30}$ & $\mathbf{6/30}$ & $\mathbf{30/30}$ & $\mathbf{23/30}$ & $\mathbf{27/30}$ & $\mathbf{21/30}$ & $\mathbf{66.1}$ \\
\bottomrule
\end{tabular}
\caption{\textbf{Real-world ablation results.} Success counts (out of 30 trials) on three real-world manipulation tasks under seen and unseen object instances, unified policy setting on Galaxea A1. \texttt{+Center}, \texttt{+Contact}, and \texttt{+Traj} denote DP3 augmented with the object center, semantic contact point, and pose trajectory, respectively. \textbf{Avg.\ SR (\%)} reports the overall success rate across all 180 trials per method. \textbf{Bold} denotes our full method.}
\label{tab:realworld_results}
\end{table}

\textbf{Implementation details.} Both the object-pose trajectory model and the action policy use a 1D conditional U-Net with FiLM conditioning at every down/mid/up block, kernel size 5, and 8 GroupNorm groups. The trajectory model uses channel widths $[64,128,256]$ and a 64-dim diffusion-step embedding; the action policy uses $[512,1024,2048]$ and a 128-dim embedding. Both models observe a history of 2 steps. The object-pose trajectory model predicts $H=4$ future poses, matching the length of the propagated affordance trajectory used for policy conditioning; it is trained for 100 steps under a squared-cosine noise schedule ($\beta\in[10^{-4},0.02]$) and sampled with DDIM for $K=50$ steps. The action policy predicts a 16-step action sequence but executes only the first 8 (action chunk length) before replanning in a receding-horizon fashion; it is trained under the same 100-step schedule and sampled with DDIM for $K=10$ steps, at a control frequency of 10\,Hz. The static affordance is encoded by a 2-layer MLP (hidden dimension 256). The propagated 4-step affordance trajectory ($\mathbb{R}^{4\times3}$) is linearly embedded, prepended with a learnable $[\mathrm{CLS}]$ token, and given sinusoidal positional encodings (length 5); a pre-norm Transformer (2 layers, 2 heads, $d_{model}=64$, feedforward dimension 256, dropout 0.1, GELU) processes the sequence, and the final $[\mathrm{CLS}]$ embedding gives the 64-dim trajectory feature used for policy conditioning. For tasks involving a single interaction target (e.g., PickCube), the source and target objects in Eq.~\eqref{eq:src_in_tgt} both refer to the same object category; the target object coordinate frame at time $t_0$ is defined by the pose estimate from FoundationPose at the first observation frame.

\subsection{Simulation Results}

We use 30 demonstrations for PickCube, StackCube, PokeCube, and PushCube, and 50 demonstrations for PegInsertion and PlugCharger.

\textbf{Baselines.}
We compare against DP3~\cite{ze20243d} (point-cloud-only imitation learning), AffordDP~\cite{wu2025afforddp} (DP3 augmented with a static contact point), and SPOT~\cite{hsu2025spotse3posetrajectory}, which conditions the policy on a predicted future SE(3) object pose trajectory.

\textbf{Analysis.}
As shown in Table~\ref{tab:sim_results} (mean\,$\pm$\,std over 5 seeds $\times$ 30 trials per seed), our approach outperforms DP3, AffordDP, and SPOT across all six tasks. The margin is widest on the precision-critical insertion tasks, PegInsertion and PlugCharger, where tracking the target's evolving pose matters most; PlugCharger's multi-contact docking requirement nonetheless keeps it the hardest task overall. We also lead on the remaining, less precision-demanding tasks, showing the benefit extends beyond high-precision settings. This advantage comes from conditioning on a time-varying geometric prior: rather than relying on instantaneous observations, the policy tracks the evolving spatial relationship between the end-effector and the target, reducing error accumulation from occlusions, sensor noise, and perceptual ambiguity.

\begin{figure}[t]
    \centering
    \includegraphics[width=0.75\linewidth]{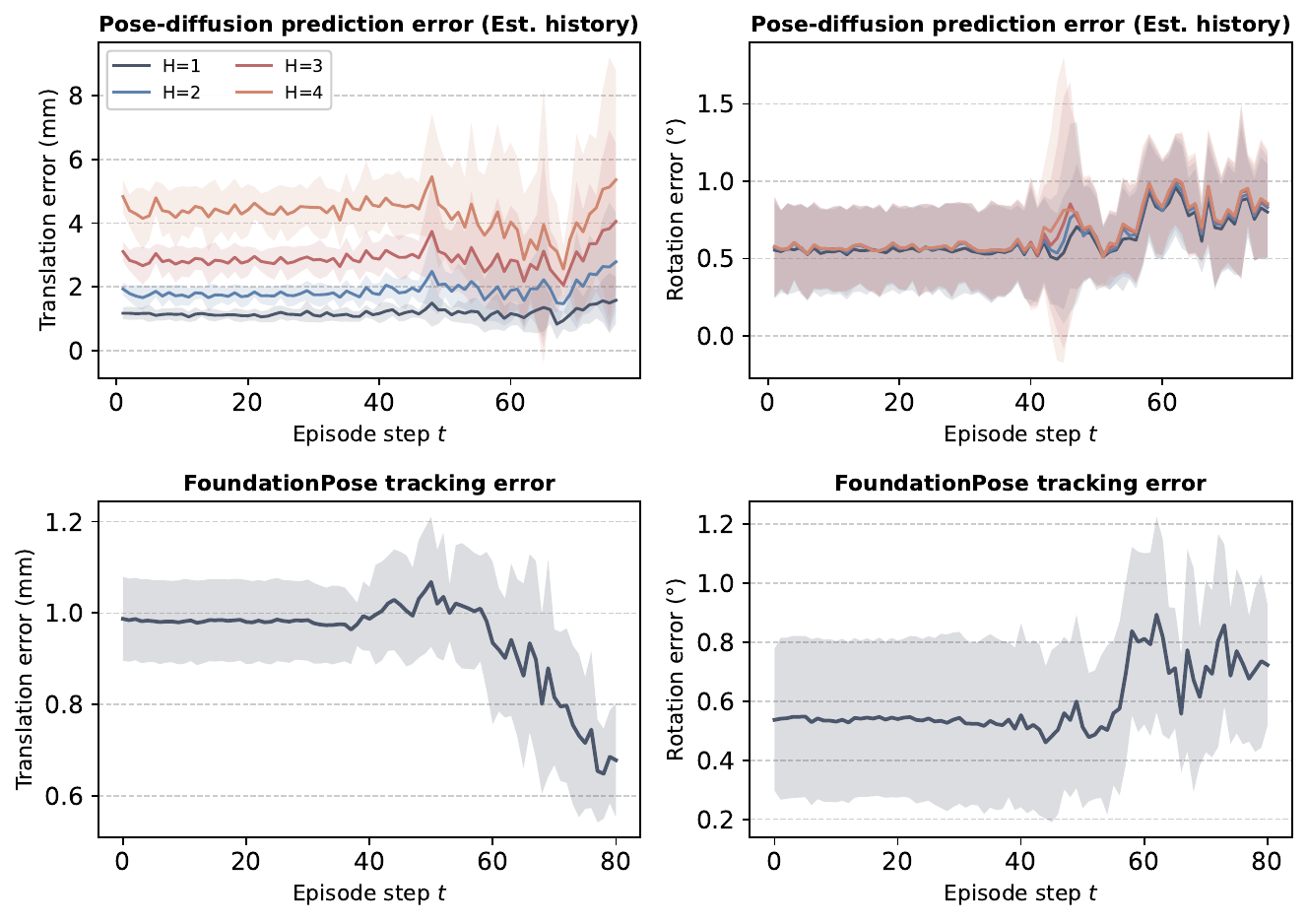}
    \caption{\textbf{Per-episode-step error analysis (StackCube, 10 held-out trajectories).} Top: pose-trajectory prediction error at each future step $k=1,\dots,4$ under FoundationPose-estimated history, evaluated at every episode step $t$. Bottom: FoundationPose's own tracking error at each episode step $t$.}
    \label{fig:error_analysis}
\end{figure}

\textbf{Pose Prediction Accuracy.} To characterize how prediction quality degrades with horizon length, we evaluate the trajectory diffusion model on held-out demonstrations, measuring translation and rotation error at each predicted step $k=1,\dots,H$ ($H=4$) under ground-truth and FoundationPose-tracked pose history. Under GT history, error grows roughly linearly with $k$, from $0.59$\,mm / $0.08^\circ$ at $k=1$ to $1.89$\,mm / $0.25^\circ$ at $k=4$; under estimated history it is consistently higher (e.g., $3.77$\,mm at $k=4$) since tracking noise propagates into the prediction, while rotation error plateaus around $1.3^\circ$ after $k=1$ rather than compounding further. Repeating this analysis at every episode step $t$ on 10 held-out StackCube trajectories (Fig.~\ref{fig:error_analysis}, top) shows the same pattern holds throughout a rollout, not just on average: error at each horizon stays roughly flat, with only a transient rotation-error spike around the contact event ($t\!\approx\!45$) that decays within a few steps, and it does not compound into growing drift over the episode. Over the same rollouts, FoundationPose's own tracking error (Fig.~\ref{fig:error_analysis}, bottom) rises and stays elevated after contact, at a magnitude comparable to or larger than the trajectory model's own prediction error. This is a correlational comparison of two separately measured error sources on a single task, not a controlled ablation, so we read it as suggestive rather than conclusive: it is consistent with upstream tracking noise being a larger contributor to post-contact pose uncertainty than compounding prediction error here, a hypothesis we discuss further and propose to test directly via noise injection in Sec.~\ref{sec:discussion}.

\subsection{Real-world Results}

On the Galaxea A1, we train a single unified policy per task on 30 teleoperated demonstrations spanning multiple object instances and diverse placements, shared across all compared methods. We test on both seen instances and unseen instances that differ in appearance and geometry, with a trial counted successful only if the task-specific goal is reached within the time horizon.

On the UR7e (Fig.~\ref{fig:setup}b), we evaluate PickCup, Ring-on-Peg, Put-in-Bowl, and USB Insertion under the same protocol against DP3, AffordDP, and SPOT (Table~\ref{tab:realworld_ur7e}). \methodname{} performs reliably on PickCup, Ring-on-Peg, and Put-in-Bowl, confirming that dynamic affordance trajectories generalize across robot arms, cameras, and task geometries. USB Insertion, which requires sub-millimeter alignment inside a narrow port, remains comparatively harder and marks the boundary of current precision.

\begin{table}[t]
\centering
\renewcommand{\arraystretch}{1.3}
\setlength{\tabcolsep}{5pt}
\begin{tabular}{l ccccc}
\toprule
\textbf{Method} & PickCup & Ring-on-Peg & Put-in-Bowl & USB Insertion & Avg.\,SR (\%)\,$\uparrow$ \\
\midrule
DP3      & 15/30 & 12/30 & 21/30 & 1/30  & 40.8 \\
AffordDP & 25/30 & 14/30 & 25/30 & 3/30  & 55.8 \\
SPOT     & 26/30 & 11/30 & 25/30 & 3/30  & 54.2 \\
\rowcolor{gray!25}
Ours     & $\mathbf{29/30}$ & $\mathbf{20/30}$ & $\mathbf{27/30}$ & $\mathbf{11/30}$ & $\mathbf{72.5}$ \\
\bottomrule
\end{tabular}
\caption{\textbf{Real-world results on UR7e.} Success counts (out of 30 trials) on four manipulation tasks, comparing against DP3, AffordDP, and SPOT baselines. \textbf{Avg.\ SR (\%)} reports the overall success rate across all 120 trials per method. \textbf{Bold} denotes our full method.}
\label{tab:realworld_ur7e}
\end{table}

\textbf{Ablation Study.}
Keeping the DP3 backbone, training data, and hyperparameters fixed, we selectively enable the semantic contact point and/or the pose trajectory, yielding five variants. DP3 uses neither, DP3+Center Point uses the object center instead of the semantic contact point, DP3+Contact Point uses the contact point only, DP3+Trajectory uses the pose trajectory only, and Ours propagates the contact point along the predicted SE(3) trajectory.

\textbf{Analysis.}
Table~\ref{tab:realworld_results} shows clear task-dependent effects. DP3+Contact Point matches or beats DP3+Center Point on five of six seen/unseen conditions, and the gap widens on unseen instances in particular (AdapterInsertion: 1/30$\to$2/30; PickCup: 0/30$\to$5/30; StackCube unseen: 8/30$\to$16/30), confirming that the benefit comes from \emph{semantic} localization of the interaction region, not just an added geometric cue. The one exception is StackCube seen, where DP3+Center Point is marginally higher (22/30 vs.\ 21/30 for DP3+Contact Point)---within the noise of a 30-trial count and consistent with the object center already being a reasonable proxy for the contact region on an unoccluded, axis-aligned cube. On PickCup and StackCube, the contact point alone already helps, and the full trajectory-propagated method extends this advantage further, most visibly when geometry and appearance both vary between seen and unseen instances. AdapterInsertion is precision-critical and highly sensitive to post-contact misalignment. Static cues (DP3, DP3+Center Point) are largely insufficient, while our dynamic affordance trajectory improves both seen and unseen success meaningfully, though unseen performance remains modest, reflecting the task's sub-millimeter precision demands rather than a failure of the underlying approach.

A notable negative result is that DP3+Trajectory \emph{degrades} below plain DP3 on StackCube unseen instances. With no semantic grounding, the object-center anchor shifts between seen and unseen instances, turning the conditioning signal into noise. This confirms that trajectory conditioning only helps when the anchor is semantically grounded. Both components are necessary for precision-demanding, unseen-instance generalization.

\section{Discussion}\label{sec:discussion}

By propagating the anchor contact point through a predicted $SE(3)$ pose trajectory, \methodname{} continuously constrains post-contact motion, improving stability under placement perturbations, especially on precision-critical, contact-rich tasks such as insertion. Our ablations suggest contact points and the affordance trajectory play complementary roles: the former localizes \emph{where} to interact, the latter guides \emph{how} to follow the interaction over time, and combining both yields the best performance. Unlike prior work that mainly strengthens visual encoders (e.g., point-cloud inputs, equivariant architectures, multi-view semantic fields), \methodname{} instead injects an explicit interaction prior maintained over time through pose-conditioned propagation---particularly valuable under limited demonstration data, where visual generalization alone cannot bridge the seen/unseen appearance gap. Our per-episode-step analysis (Sec.~\ref{sec:experiments}) further shows that, on the tasks and rollouts we test, pose-trajectory prediction error does not compound into growing drift over the course of an episode. Whether this non-accumulation property also holds under more adverse tracking conditions, such as heavy occlusion or injected pose noise, and whether it yields a measurable robustness advantage over one-shot baselines such as AffordDP under matched perturbations, remains to be directly validated; the evidence we present is a correlational, single-task comparison rather than a controlled ablation, and we treat the stronger claim as a hypothesis for future work rather than an established result.

\section{Conclusion}\label{sec:conclusion}

We identified a gap in existing affordance-guided policies: conditioning signals either preserve \emph{where} to contact (semantic affordances) or \emph{how} that contact evolves (object pose trajectories), but not both. We proposed \methodname{}, which propagates a retrieved anchor affordance through a predicted SE(3) pose trajectory to obtain a dynamic affordance trajectory, giving the policy state-consistent, time-varying guidance. Across six ManiSkill3 tasks and seven real-world tasks on two robot platforms, \methodname{} consistently outperforms DP3, AffordDP, and SPOT, with the largest gains on precision-critical, contact-rich tasks, and controlled ablations show that semantic contact retrieval and pose-trajectory propagation are each necessary but individually insufficient. We see this as a step toward affordance representations that remain valid throughout an interaction rather than only at first contact, with richer contact semantics, uncertainty-aware pose prediction, and tactile or force feedback as natural directions for extending this to a wider range of manipulation tasks.

\bibliographystyle{plain}
\bibliography{references}

\end{document}